# Detection of high-frequency oscillations using time-frequency analysis

**Mostafa Mohammadpour**[1+]
**Mehdi Zekriyapanah Gashti**[2,3]
**Yusif S. Gasimov**[4,5,6]

[1]*Department of Computational Perception, Johannes Kepler University, Linz, Austria.*
Email: neuroengineering.research@gmail.com
[2]*Department of Computer Engineering, University of Applied Science and Technology, Tehran, Iran.*
[3]*Department of Computer Engineering, Payame Noor University, Tehran, Iran.*
Email: gashti@pnu.ac.ir
[4]*Azerbaijan University, Baku, Azerbaijan.*
[5]*Institute of Mathematics and Mechanics, Baku, Azerbaijan.*
[6]*Institute for Physical Problems, Baku State University, Baku, Azerbaijan.*
Email: yusif.gasimov@au.edu.az

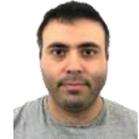

*(+ Corresponding author)*

## ABSTRACT

**Article History**
Received: 9 May 2025
Revised: 3 July 2025
Accepted: 7 August 2025
Published: 25 August 2025

High-frequency oscillations (HFOs) are a new biomarker for identifying the epileptogenic zone. Mapping HFO-generating regions can improve the precision of resection sites in patients with refractory epilepsy. However, detecting HFOs remains challenging, and their clinical features are not yet fully defined. Visual identification of HFOs is time-consuming, labor-intensive, and subjective. As a result, developing automated methods to detect HFOs is critical for research and clinical use. In this study, we developed a novel method for detecting HFOs in the ripple and fast ripple frequency bands (80-500 Hz). We validated it using both controlled datasets and data from epilepsy patients. Our method employs an unsupervised clustering technique to categorize events extracted from the time-frequency domain using the S-transform. The proposed detector differentiates HFO events from spikes, background activity, and artifacts. Compared to existing detectors, our method achieved a sensitivity of 97.67%, a precision of 98.57%, and an F-score of 97.78% on the controlled dataset. In epilepsy patients, our results showed a stronger correlation with surgical outcomes, with a ratio of 0.73 between HFO rates in resected versus non-resected contacts. The study confirmed previous findings that HFOs are promising biomarkers of epileptogenicity in epileptic patients. Removing HFOs, especially fast ripples, leads to seizure freedom, while remaining HFOs lead to seizure recurrence.

**Contribution/Originality:** This study introduces a novel unsupervised clustering method for automatic HFO detection using time-frequency features from the S-transform, enabling differentiation between true HFOs, spikes, and artifacts. The proposed approach outperforms existing detectors in both sensitivity and clinical relevance, demonstrating a strong correlation between detected HFOs and surgical outcomes in epilepsy patients.

## 1. INTRODUCTION

Epilepsy is defined as a neurological disorder that affects 1-2% of the world's population. Epileptic seizures are characterized by neural discharges resulting from abnormal brain activity in the cerebral cortex. Many patients with epilepsy can be treated with anti-epileptic medications. For those resistant to medication, epilepsy surgery is a therapeutic option to achieve seizure freedom. The epileptogenic zone in these patients is delineated by the seizure onset zone (SOZ), which is considered the gold standard and the most widely used marker, and can be identified through presurgical evaluation of intracranial electroencephalography (iEEG) data.

High-frequency activities serve as biomarkers in the detection of epileptogenic tissue and are essential markers for cortical function processing in the motor and sensory cortex. HFOs are categorized into high-gamma (65-100 Hz), ripple (80-250 Hz), and fast ripple (250-500 Hz) [1], which are involved in task-induced activity and seizure generation, with notable spatial and temporal resolution. High-gamma and ripple oscillations generally occur in the neocortex, while fast ripples are observed in the hippocampus [2]. HFOs can also be recorded during the inter-ictal state of an epileptic seizure, at rest, or in a task-induced paradigm. They should have at least six peaks greater than three standard deviations to be clearly distinguished from background noise [3]. The precise characteristics and definition of clinically associated HFOs are still not well-defined in clinical literature.

Visual marking of HFO events provides a clear relationship between epilepsy and the epileptogenic zone [4]. Most clinical experts usually perform initial seizure analysis visually, which is not suitable for long-term monitoring. HFO marking could become a critical issue if there is no formal definition of HFOs and no clear morphology. However, visual marking of HFOs is a cumbersome, highly time-consuming, and subjective task. Therefore, developing methods for automatically detecting HFOs is essential for clinical practice and research.

Most existing HFO detection algorithms perform detection in two steps: transforming the signal into a specific domain to extract features or filtering the signal into a specific range, and classifying the features into desired categories (for example, HFO, background, spike, and artifact). Feature extraction can be accomplished in the time, frequency, or both time and frequency domains. Methods that utilize the time-frequency domain generally outperform others because of the non-stationary characteristics of brain signals.

The differences among various detection algorithms lie in computing the energy function of the filtered signals. Some studies use root mean square (RMS) to differentiate HFOs from background activities, while others use signal line length (SLL) or the Hilbert transform, which introduced the first HFO detection method [3]. The method filters the signal in the 80-500 Hz range and then applies a moving average RMS to extract some events of interest (EOI). Additionally, some EOIs remain after applying the threshold. Study Gardner et al. [5] proposed an HFO detection method called short line length (SLL), which computes the line length energy of a sliding window from a bandpass-filtered signal with lower frequency cutoffs (30-85 Hz). The method retained events if their amplitude was greater than the 97.5th percentile of the empirical cumulative distribution function (CDF) for each epoch. Crépon et al. [6] showed the importance of the Hilbert transform for computing the energy of the filtered signal. Zelmann et al. [7] proposed a method called the Montreal Neurological Institute (MNI) detector. It combines three detectors for identifying EOIs, a baseline detector, and two HFO detectors. Fedele et al. [8] demonstrated the effectiveness of time-frequency analysis (TFA) in HFO detection. First, they used a baseline detector based on time-frequency entropy to eliminate artifacts using the Stockwell transform. Then, they differentiate real EOIs from spurious ones in the time-frequency domain using isolated energy contribution, which indicates salience over background activity. Finally, they utilized information from multichannel signals to remove artifacts in the remaining events from the previous step. Burnos et al. [9] developed an HFO detection algorithm to classify events into four categories based on their morphology. They defined four types of HFO classes according to the waveform's amplitude and frequency. Subsequently, they investigated the relationship between HFOs, the seizure onset zone (SOZ), resection area, and surgical outcomes. They found that the rate of HFOs within the SOZ was significantly higher than outside it. Liu et al. [10] hypothesized that pathological high-frequency oscillations (HFOs) have a similar waveform in the seizure onset zone (SOZ) and investigated the pattern of this waveform using an automatic HFO detection method in 18 subjects (13 with epilepsy and 5 controls). In their method, a bandpass filter in the 80-500 Hz range is applied to raw signals, followed by an amplitude-based detector to eliminate background activity. Next, the Short-time Fourier Transform (STFT) is applied to the raw signal in each retained candidate. Entropy of the spectrum, sub-band power ratio, and the frequency corresponding to the maximum peak-to-notch ratio are computed as features for input into a clustering algorithm. Migliorelli et al. [11] proposed an automated detector using S-Transform time-frequency distribution and Gaussian mixture clustering to identify HFO events. They utilized time-frequency features, such as entropy, area, time, and frequency width of extracted events, to distinguish significant event activities for HFOs and non-HFOs. Visualizing the iEEG signal for HFO analysis is highly time-consuming and subjective. Therefore, automatic HFO detection is necessary for clinical applications to localize SOZ areas in the presurgical evaluation step. We are also investigating whether resection of the HFO areas leads to seizure freedom in epileptic patients. To achieve this, we developed a new method for detecting HFOs, called time-frequency event clustering (TFEC), which detects events in ripple and fast ripple bands (80-500 Hz) based on time-frequency analysis methods. We used different processing pipelines for detecting HFOs compared to Migliorelli et al. [11]. Features in the time, frequency, and time-frequency domains were utilized, along with a feature selection algorithm to identify the best feature sets. An unsupervised clustering technique was employed to classify events into HFO and non-HFO categories. The performance of the proposed method is evaluated against two different datasets: the controlled dataset [12] and epileptic patients' data with the surgery outcome [13].

## 2. METHODS

### 2.1. Simulated and Epileptic Patients Data

To assess detected HFOs against ground truth, we used publicly available stereo-EEG (sEEG) data recorded during non-REM slow-wave sleep in epileptic patients [12]. Data consists of 8 channels, 2 minutes of recording length, and a sampling frequency of 2048 Hz. It also includes 30 different backgrounds for each channel across 4 different SNRs (0, 5, 10, and 15 dB), totaling 960 different combinations. The data contains various events: ripple, fast ripple, and spike, which can co-occur, such as a ripple occurring simultaneously with a fast ripple.

In another experiment, we used data from epileptic patients, which includes the surgery outcome scale, to assess the relationship between the HFO rate and surgical success. To do this, we utilized publicly available data [13] that includes 20 drug-resistant epileptic patients with focal epilepsy. These patients had both sEEG and subdural grid and strip electrodes implanted. Data were recorded at a sampling frequency of 2000 Hz, and patients were followed for more than a year after surgery to determine seizure freedom. The seizure outcomes were graded according to the International League Against Epilepsy (ILAE) scale, where class 1 indicates complete seizure freedom, and class 6 indicates a 100% increase over baseline.

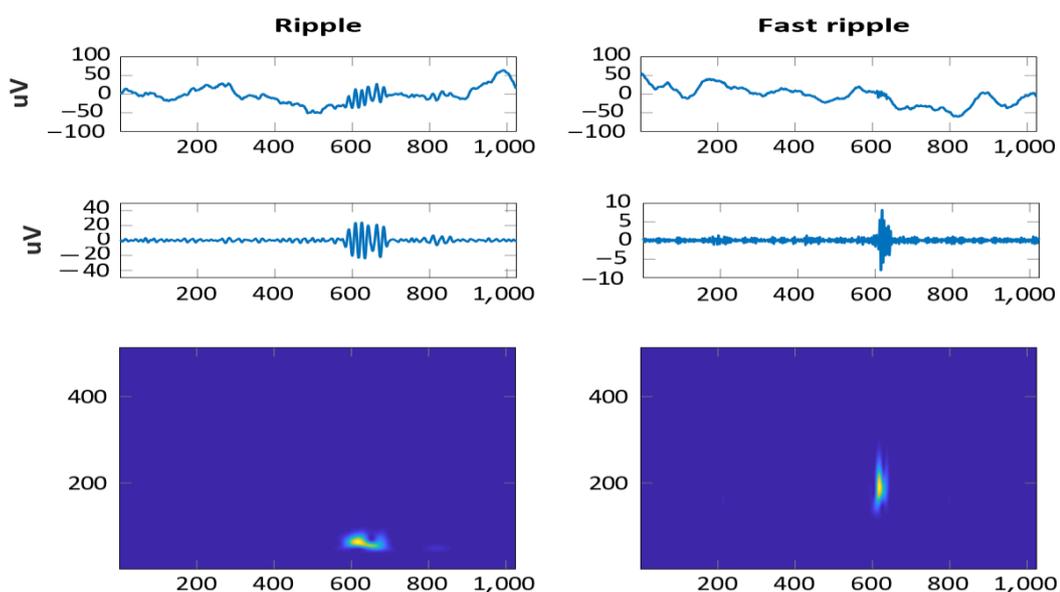

**Figure 1.** Example of HFO events in different frequency bands, including ripple and fast-ripple. The first row displays raw signals, the second row shows filtered signals, and the third row presents a time-frequency map of signals using the S-transform, where HFOs appear as islands or blobs.

*2.2. Automatic Detection of HFOs*

Time-frequency distributions (TFDs) are central to this work for extracting EOIs from iEEG signals because they can extract useful information from non-stationary signals. HFOs are short-lived events that appear as islands or isolated blobs in the TFD map (Figure 1), while they cannot be easily recognized solely in the time or frequency domains. Algorithms operating only in the time or frequency domain are usually inadequate for processing non-stationary signals like EEG, leading to poor detection accuracy.

Figure 2 illustrates the HFO detection pipeline, which contains a three-stage pattern recognition approach. The algorithm consists of three major parts: extracting events from the TFD representation and creating pooled signals from time points corresponding to the events, extracting features from pooled signals, and clustering the signals into HFO or non-HFO events.

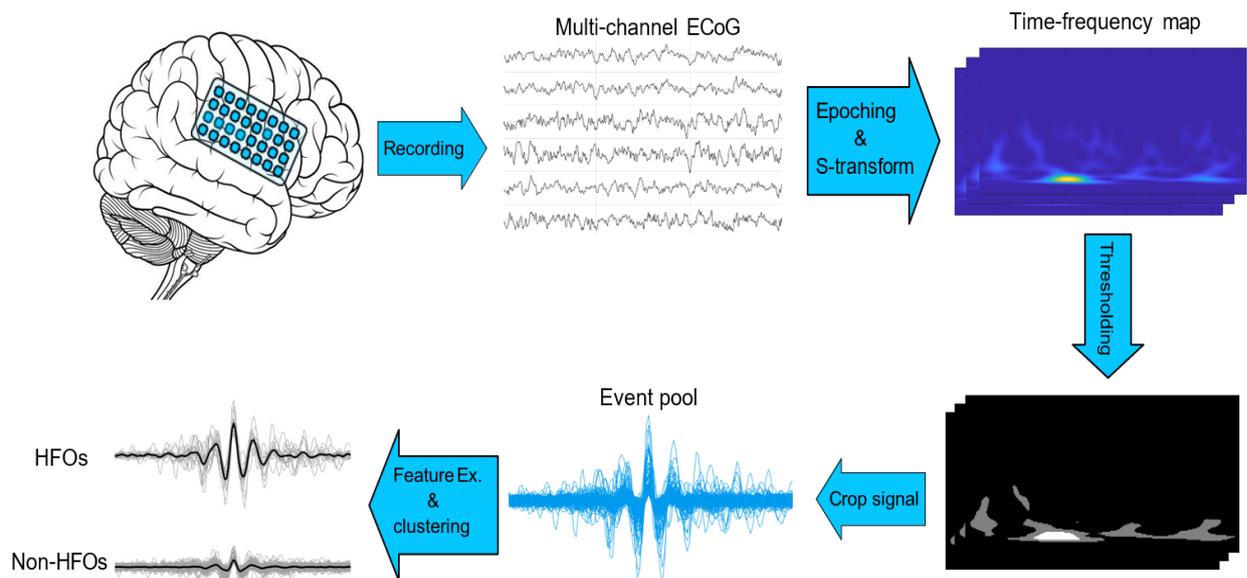

**Figure 2.** HFOs detection workflow. After signal epoching, EOIs are extracted from the representation of TFDs, and signals are cropped based on the time points of events. Next, features are extracted from pooled events. Finally, a clustering algorithm is applied to detect HFOs from non-HFO events.

*2.2.1. First Stage: Extract Events*

At the beginning, signals are epoched with a 1-second window and 0.5-second overlapping to extract events from the continuous EEG signal. Then, epoch signals are denoised using the Teager-Kaiser energy operator (TKEO). The TKEO is used to improve the signal-to-noise ratio (SNR) and minimize the error in electromyography (EMG) onset detection. The discrete form of TKEO is defined as.

$$TKEO\{x[n]\} = x2[n] - x[n-1]x[n+1] \qquad (1)$$

Where $x$ is the iEEG signal value and $n$ is the sample number. TKEO is applied after the signal is bandpass filtered. It provides perfect time resolution because it requires only three samples for energy computation at each time instant. Next, EOIs should be extracted from TFDs.

An EOI is a rising event or high-frequency peak over background activities with concentrated energy in the time-frequency domain. The time-frequency representation was acquired using the S-transform, demonstrating its advantages in HFO detection [8]. S-transform of a signal is an extended version of the short-time Fourier transform (STFT), which addresses the cross-term problem [14]. The S-transform of the given signal $s(t)$ is denoted by $S(t, f)$ and is defined as the CWT of the signal multiplied by a phase factor of frequency $f$.

$$S_x(t,f) = \int_{-\infty}^{\infty} s(\tau) \frac{|f|}{\sqrt{2\pi}} e^{-\pi(t-\tau)^2 f^2} e^{-j2\pi ft} dt \qquad (2)$$

S-transforms have better frequency resolution in the low-frequency range if their window size is large in the time domain, and they have better time resolution in the high-frequency range if their window size is narrow. After calculating TFDs, EOIs should be extracted from the map. Traditional image binarization uses a threshold value to convert an image pixel to a binary map.

There is a drawback to using a fixed threshold, which is inconsistent with different signal properties in different channels because there is high variation in the intensity of the TFD map across different channels. Therefore, we need a robust image binarization method with no threshold parameters that easily adapts to various EEG signals. Here, we used the Otsu thresholding method, a type of automatic image thresholding that can identify blobs in the TFD image map [15].

The Otsu method is an adaptive thresholding technique that iterates over all possible threshold values of a given image to minimize within-class variance or, equivalently, maximize between-class variance.

By having island or blob objects in the TFDs map, we need to measure their region properties for further processing. The connected-component labeling (CCL) algorithm is applied to binary images to identify objects. The image features, such as area, centroid, width, and height of each object, are calculated at the cropping time point of the corresponding blob in the signal, as shown in Figure 3.

TFDs contain a wealth of information that cannot be used for all points as features for feeding into a classification or clustering algorithm because it increases the feature space's dimensionality. To avoid this curse of dimensionality, a subset of features must be extracted that represent relevant information about the events of interest. We calculated the time point corresponding to the object center on the TFDs map with a confidence interval of 200 ms. Due to signal epoching, some duplicated events in the event pool should be removed. Then, events are aligned for further processing, including feature extraction and clustering.

### 2.2.2. Second Stage: Feature Extraction

All extracted events from the first stage underwent a feature extraction process to identify signal characteristics for the clustering algorithm. To accomplish this, we used three groups of signal features in the time, frequency, and time-frequency domains.

Time domain features: Several time-domain features are directly related to the statistical properties of a signal under the assumption that normal and abnormal signals have different probability distributions, implying that the parameters characterizing the probability distributions of normal and abnormal signals can be used as features. A list of statistical features that can be extracted from the time-domain representation is provided in Table 1. The remaining time-domain features are listed in Table 2.

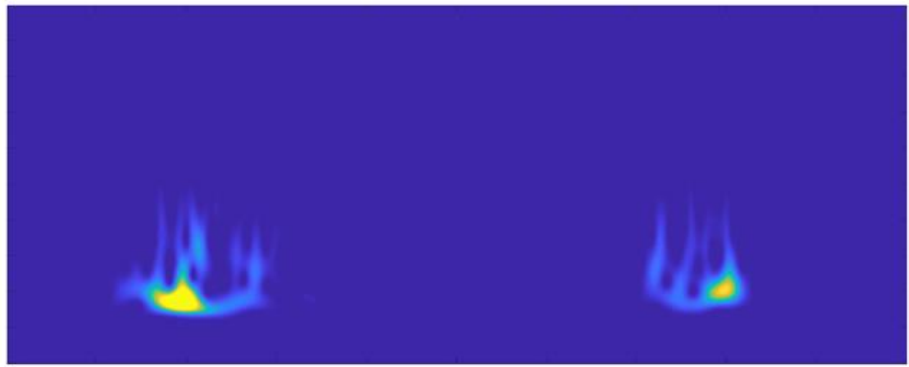

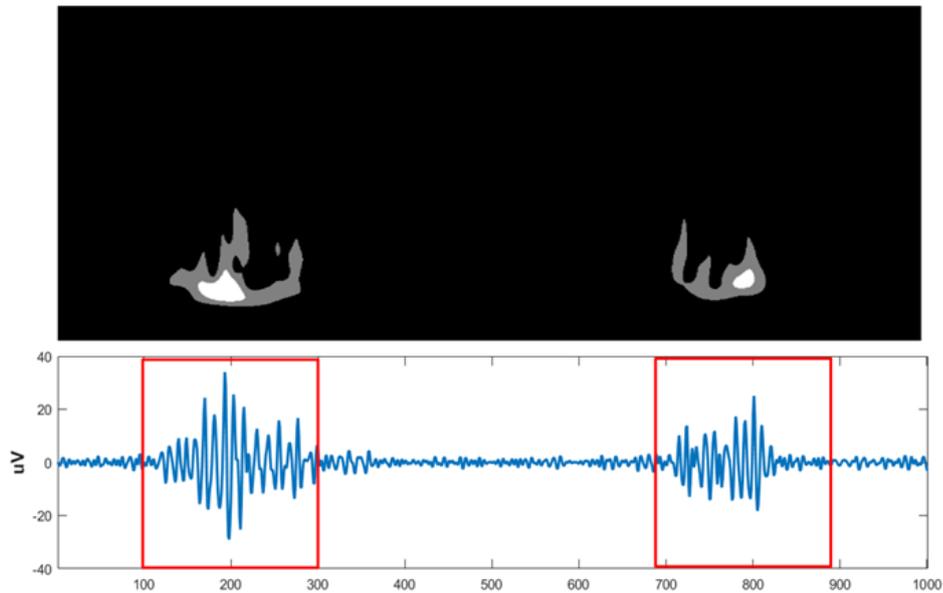

**Figure 3.** Extract EOIs from the TFDs map, then crop the signal corresponding to the events.

**Table 1.** Statistical Time-Domain Features.

| Feature | Formula |
|---|---|
| Mean | $m_{(t)} = \frac{1}{N} \sum_n x[n]$ |
| Variance | $\sigma^2_{(t)} = \frac{1}{N} \sum_n (x[n] - m_{(t)})^2$ |
| Skewness | $\gamma_{(t)} = \frac{1}{N} \sum_n (x[n] - m_{(t)})^3$ |
| Kurtosis | $k_{(t)} = \frac{1}{N} \sum_n (x[n] - m_{(t)})^4$ |
| Coefficient of variance (CV) | $c_{(t)} \frac{\sigma_{(t)}}{m_{(t)}}$ |

**Table 2.** Other time-domain features.

| Feature | Formula |
|---|---|
| Root mean square (RMS) | $RMS = \sqrt{\left(\frac{1}{n}\right) \sum_{i=1}^{n} (x_i)^2}$ |
| Power | $P(x) = \sum_{i=1}^{N-1} \|x[i]\|^2$ |
| Line length | $P(x) = \sum_{i=1}^{N-1} \|x[i]\|^2$ |
| Autocorrelation function | $r_k = \frac{\sum_{i=1}^{N-k} (Y_i - \bar{Y})(Y_{i+k} - \bar{Y})}{\sum_{i=1}^{N} (Y_i - \bar{Y})^2}$ |
| Nonlinear energy | $E(x) = \sum_{i=1}^{N-2} (x^2[i] - x[i+1]x[i-1])$ |
| Fractal dimension | $D_B(X) = \lim_{\epsilon \to 0} \frac{\log N(\epsilon)}{\log (1/\epsilon)}$ |
| Signal range | $(x) = \|max(x) - min(x)\|$ |

Frequency domain features are used to detect abnormalities in EEG signals [16]. Power spectral density (PSD) is commonly used for extracting frequency information from a given signal and is robust for identifying spectral patterns of a signal.

Here, we first compute the PSD of a signal using the Welch method, and then statistical parameters and other features are extracted as shown in Table 3. The PSD is calculated by a Hamming window of 1 second without an overlapping window.

**Table 3.** Frequency-domain features.

| Feature | Formula |
| --- | --- |
| Spectral flux | $\mathcal{FL}_{(f)} = \sum_{k=1}^{M/2} |Z_r[k] - Z_{r-1}[k]|$ |
| Spectral flatness | $S\mathcal{F}_{(f)} = M \frac{(\prod_{k=1}^{M} |Z_x[k]|)^{M^{-1}}}{\sum_{k=1}^{M} |Z_x[k]|}$ |
| Spectral entropy | $SE(X) = -\sum_{k} x[k] \log x[k]$ |
| Intensity-weighted mean frequency (IWMF) | $IWMF(X) = \sum_{k} x[k]f[k]$ |
| Intensity-weighted bandwidth (IWBW) | $IWBW(X) = \sqrt{\sum_{k} x[k]f[k] - IWMF(X)^2}$ |
| Peak power | $PP = max(pxx$ |

Time-frequency domain features: High-frequency activities have different probability distribution functions (PDFs) in the TFDs, which can be characterized by various feature sets, including statistical parameters, entropy features, image features, and other TFD-related parameters. The statistical parameters of a given signal s(t) in (t, f) can be calculated as shown in Table 4. Here, time-domain features are simply extended to joint time-frequency features by replacing time-domain moments with corresponding time-frequency moments. The remaining time-frequency domain features are listed in Table 5.

**Table 4.** Statistical features in the time-frequency domain.

| Feature | Formula |
| --- | --- |
| Mean | $m_{(t,f)} \& = \frac{1}{NM} \sum_{n=1}^{N} \sum_{m=1}^{M} \rho_z[n,m]$ |
| Variance | $\sigma^2_{(t,f)} \& = \frac{1}{NM} \sum_{n=1}^{N} \sum_{m=1}^{M} (\rho_z[n,m] - m_{(t,f)})^2$ |
| Skewness | $\gamma_{(t,f)} \& = \frac{1}{NM\sigma^3_{(t,f)}} \sum_{n=1}^{N} \sum_{m=1}^{M} (\rho_z[n,m] - m_{(t,f)})^3$ |
| Kurtosis | $k_{(t,f)} \& = \frac{1}{NM\sigma^4_{(t,f)}} \sum_{n=1}^{N} \sum_{m=1}^{M} (\rho_z[n,m] - m_{(t,f)})^4$ |
| Coefficient of variation (CV) | $c_{(t,f)} \& = \frac{\sigma_{(t,f)}}{m_{(t,f)}}$ |

**Table 5.** Other time-frequency domain features.

| Feature | Formula |
|---|---|
| Shannon entropy | $ShE_{(t,f)} = -\sum_{n=1}^{N} \sum_{m=1}^{M} (\rho_r[n,m]) \log_2 |\rho_r[n,m]|$ |
| Renyi entropy | $RE_{(t,f)} = \frac{1}{1-\alpha} \log_2 \sum_{n=1}^{N} \sum_{m=1}^{M} (\rho_r[n,m])^\alpha$ |
| Time-frequency flatness | $SF_{(t,f)} = MN \frac{\prod_{n=1}^{N} \prod_{m=1}^{M} |\rho_z[n,m]|^{\frac{1}{NM}}}{\sum_{n=1}^{N} \sum_{m=1}^{M} \rho_z[n,m]}$ |
| Time-frequency flux | $\mathcal{FL}_{(t,f)} = \frac{1}{NM} \sum_{n=1}^{N-l} \sum_{m=1}^{M-k} |\rho_z[n+l, m+k] - \rho_z[n,m]|$ |
| Energy concentration | $Con_{(t,f)} = \left( \sum_{n=1}^{N} \sum_{k=1}^{M} |\rho[n,k]|^{\frac{1}{2}} \right)^2$ |

TFD image-based features: The TFD representation of a signal can be considered a grayscale image. Therefore, image processing techniques are valuable tools for extracting features from TFD maps. These include geometrical features and local binary pattern (LBP) [17]. Meaningful regions in an image, also known as regions of interest (ROI), are defined areas within an image that contain a concentration of energy with specific patterns that are easily distinguishable visually. Here, image moment features are used to characterize the ROIs of an image. To compute image moments, an image must be transformed into a binary image by applying an appropriate threshold. We used the Otsu method for image binarization, which was also employed in the event extraction stage.

The moments from a segmented binary image, including centroid, height, width, compactness, area, and perimeter, help describe image segment features.

For a given grayscale image $\rho_z[n, m]$ with L number of segments, the moment of order (p, q) can be extracted as follows:

$$m_{pq}^l = \sum_m \sum_n m^p n^q \rho_z^l[n,m] \quad (3)$$

Where $p, q = 0, 1, 2, ...$, and $\rho_z^l[n, m]$ is the *lth* segment of the given image. **Central** moments of an image are defined as follows:

$$\mu_{pq}^l = \sum_m \sum_n \left(m - \overline{m^l}\right)^p \left(n - \overline{n^l}\right)^q \rho_z^l[n,m] \quad (4)$$

Where are the components of the centroid for the *lth* segment?

Other geometric features of a TFDs image, including area, perimeter, and compactness, are estimated as follows:

- The area: $\mu_{00}^l$
- The perimeter: $(m_{30}^l + m_{12}^l)^2 + (m_{03}^l + m_{21}^l)^2$
- The compactness: $((m_{30}^l + m_{12}^l)^2 + (m_{03}^l + m_{21}^l)^2)^2 / \mu_{00}^l$

Local binary patterns: describe the local texture features of a given image. This method considers the 8-neighborhood of image pixels and the threshold neighborhood of each pixel; then it transforms each value from binary to a decimal value to assign a label to the central pixel. After applying the LBP operator to a given image and performing a histogram on its results, we can extract features by calculating different statistical parameters such as mean, variance, skewness, and kurtosis.

### 2.2.3. Third Stage: Event Clustering

Clustering is a popular form of unsupervised learning that separates data into different homogeneous subgroups based on distances between data points [18]. Points within each subgroup are more similar to each other than to points in different groups. Due to the variability in EEG data characteristics, training a classifier and applying it to other datasets can be challenging. Support Vector Machines (SVMs) are powerful supervised learning algorithms widely used in various classification and regression tasks because of their ability to handle high-dimensional data and identify optimal decision boundaries [17]. Here, we hypothesize that HFO events can be distinguished from other events, such as spikes, noise, and artifacts, using clustering because they are prominent in the TFD maps, appearing as islands or isolated blobs. Due to the high variability in the morphology of TFDs, it is preferable to design a detection system that categorizes TFDs into a known number of groups. Therefore, clustering is used for HFO detection rather than classification. A hierarchical clustering algorithm is employed to separate events into two groups: HFO and non-HFO. This tree-based clustering method builds nested clusters by successively merging or splitting them. To use clustering as a detector, clustering results specifically, signal events are averaged (as shown in Figure 2), and signals with large amplitude ranges are considered HFO, while minor signals are classified as non-HFO events or background activities.

### 2.3. Performance Evaluation

Two different datasets were used for validation to assess the proposed method. The first one includes ground truth data to estimate the sensitivity, precision, and F-score of the detection algorithm, and the second one includes surgical outcomes to find a correlation between HFOs and seizure freedom.

Compare with labeled data: The simulated dataset consists of labeled events marked by experts and is considered the gold standard. We used a time window of 100 ms in the center of each event as a confidence interval (CI) for assessing the algorithm. CIs that overlap with HFO events are considered true positives (TP), and CIs without overlap with HFO events are considered false negatives (FN). Detected HFO events outside the CIs are false positives (FP). We do not count true negatives (TN) because they are not clearly defined in the context of HFO events. The performance of the detector is evaluated based on sensitivity, precision, and F-score as follows:

$$Sens = TP/(TP + FN)$$
$$Prec = TP/(TP + FP)$$
$$F - score = 2 \times (Prec \times Sens)/(Prec + Sens) \qquad (5)$$

Sensitivity measures the detector's ability to identify true HFOs, while precision assesses the detector's ability to reject events that are not actual HFOs. The F-score, which is the harmonic mean of sensitivity and precision, serves as a measure of accuracy. Compare with surgical outcomes: The correlation of HFO rate to resection area and surgical outcomes is calculated to extend the proposed detector's clinical application. We are investigating whether high rates of HFOs can be considered a biomarker for epileptogenic areas and whether the majority of removed HFOs during surgery will result in seizure freedom in epileptic patients.

First, we measure the HFO rate in resected and non-resected contacts. Then, a ratio is defined for the rate of HFOs in resected areas and non-resected electrodes. Using this ratio, a value indicating +1 points out that the majority of HFOs are resected in patients, while a value of -1 denotes that most HFOs remain untouched. The ratio between the rates of HFOs in resected electrodes is as follows:

$$\text{Ratio Rate } (e) = \frac{\sum_{RA} \text{Rate }_e - \sum_{NonRA} \text{Rate }_e}{\sum_{[RA,NonRA]} \text{Rate }_e} \qquad (6)$$

Where $e$ is the type of event (ripple or fast ripple), RA are the resected channels, and Non-RA are the channels that were not resected.

*2.4. Parameter Optimization*

Data becomes sparse due to numerous feature sets, and the curse of dimensionality poses challenges in data analysis. To reduce model complexity, parameter optimization is necessary to identify the best feature sets that enhance model accuracy. Clustering algorithms typically depend on different distance measures, which group points close to each other into one cluster, while points far apart are assigned to different clusters. However, distance measures are less effective in high-dimensional spaces; therefore, reducing the feature space can lead to more effective distance measures in clustering algorithms. Dimensionality reduction can be achieved through two different methods: feature selection and feature projection, such as Principal Component Analysis (PCA). While PCA identifies the optimal linear transformation of the given features and reduces their number, it may not be suitable for interpreting features as biomarkers because PCA results are combinations of all input features (i.e., projecting into new feature sets) and are not easily interpretable. Feature selection methods are divided into two groups: filter and wrapper methods. The filter method uses statistical ranking of features to determine their importance. However, wrapper methods combine given features and apply an algorithm to find the best accuracy. With a set of n features, wrapper methods select a subset of d features (d < n) that maximize detector performance. Here, we used the F-score value as the cost function for parameter optimization.

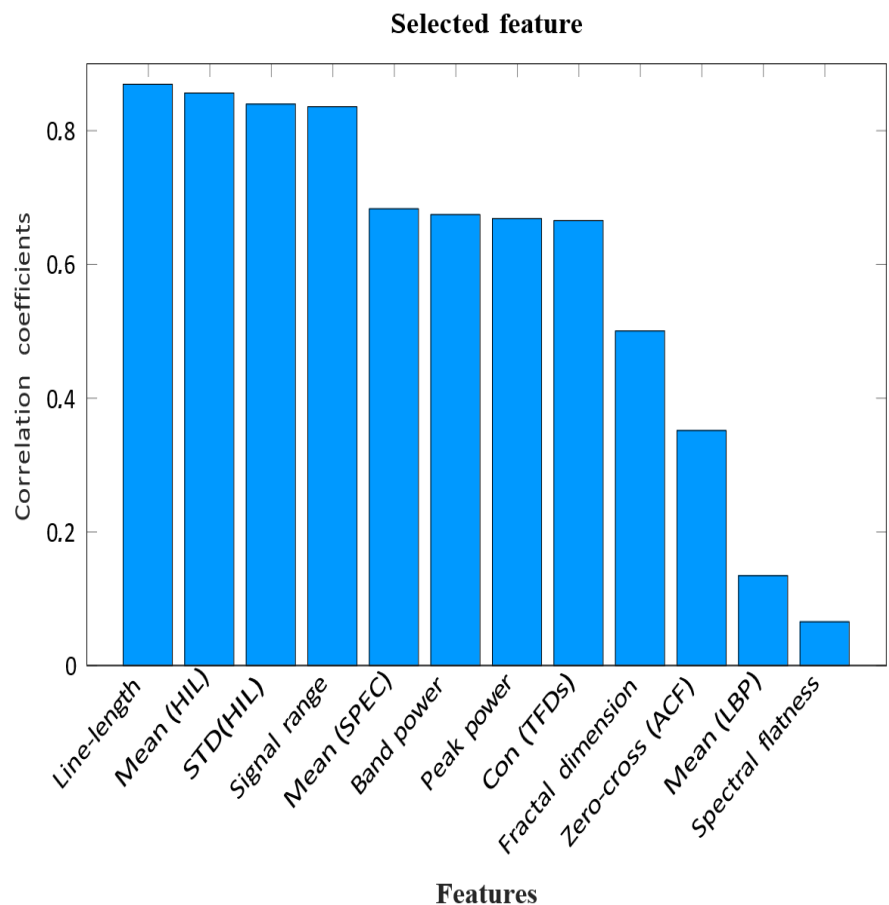

**Figure 4.** Correlation coefficients of selected features to true labels.

Wrapper methods attempt to use a subset of features to train a model and validate it against ground truth to identify the most accurate feature set. In this process, the selected features are input into a clustering algorithm, and the F-score is calculated as a cost function for the wrapper method. The sequential forward floating selection (SFFS) algorithm is employed for feature selection. The optimal accuracy was achieved by selecting 12 features, as shown in Figure 4.

## 3. RESULTS

In this study, the proposed HFO detection method is validated on both simulated data and data from epileptic patients. The controlled dataset is selected to identify the most effective feature sets for HFO detection and to evaluate its performance. The data from epileptic patients are used to correlate HFOs with patient seizure freedom after surgery. If we only compare the detector to surgical outcomes, then we do not have an HFO detector; we have an outcome detector. Therefore, an effective approach is to assess the algorithm across different scenarios: compare it with labeled data from the controlled dataset and with surgical outcome scores from epileptic patients.

### 3.1. Simulated Signal

The detector's performance is validated on a public dataset [19], which includes labeled data for different events: ripple, fast ripple, and spike. Because measuring the performance of an HFO detector is challenging, it is preferable to work in a controlled environment where the time and location of an HFO event are known in advance. This dataset includes signals with varying SNRs from 0 to 15 dB, with 30 backgrounds across 8 sEEG channels.

Figure 5 compares the results obtained from our method to those of detectors Roehri et al. [12] and Burnos et al. [20] on public datasets in terms of sensitivity, precision, and F-score for two different frequency ranges: ripple and fast ripple.

As seen in Figure 6 , there is an expectation of low sensitivity at low SNR 0 dB because HFOs amplitude is close to background activity. From the figure, it can be observed that the performance of all detectors increases as SNR increases. At SNR 15 dB, with a higher amplitude of HFOs than background activity, the detectors reach their optimal performance.

Our method's low SNR performance compared to Delphos' method is because Delphos uses different background denoising during the pre-processing stage, which results in better performance. Our method is comparable to Delphos' detector in terms of higher SNR rate, and it performs better than it in terms of F-score, which combines sensitivity and precision.

For comparing methods across different SNRs, a permutation test with 10,000 permutations was applied to each HFO label to randomly change the ripple, fast-ripple, and spike labels. Our method detects HFOs in the ripple band with an F-score of 25%, a confidence interval (CI) of [14-37%], and a p-value of 0.49 for SNR 0; 73%, CI [55-86%], p=0.012 for SNR 5; 93%, CI [88-97%], p=0.010 for SNR 10; and 95%, CI [92-98%], p=0.009 for SNR 15. It also detects HFOs in the fast ripple band with an F-score of 16%, CI [13-18%], p=0.43 for SNR 0; 90%, CI [85-94%], p=0.010 for SNR 5; 98%, CI [97-99%], p=0.011 for SNR 10; and 99%, CI [97-100%], p=0.009 for SNR 15.

### 3.2. Epileptic Patients Data

The ratio of HFO rate between the resected area and the non-resected area should be high for patients with good outcomes compared to those with poor outcomes. We used the ratio metrics explained in section 2.3 to assess the HFO rates.

Ripple and fast ripple rates are calculated for each electrode. SNRs for ripple and ripple background are 26.03 dB and 13.33 dB, respectively, and for FR and FR background are 5.30 dB and 1.22 dB, respectively. Figure 6 shows the ratio of HFO rates across all patients with different HFO detection methods. The ratio between HFO rates in resected and non-resected electrodes is significantly higher in patients with good outcomes (ILAE 1) than in those with poor outcomes (ILAE 2-6) using our method.

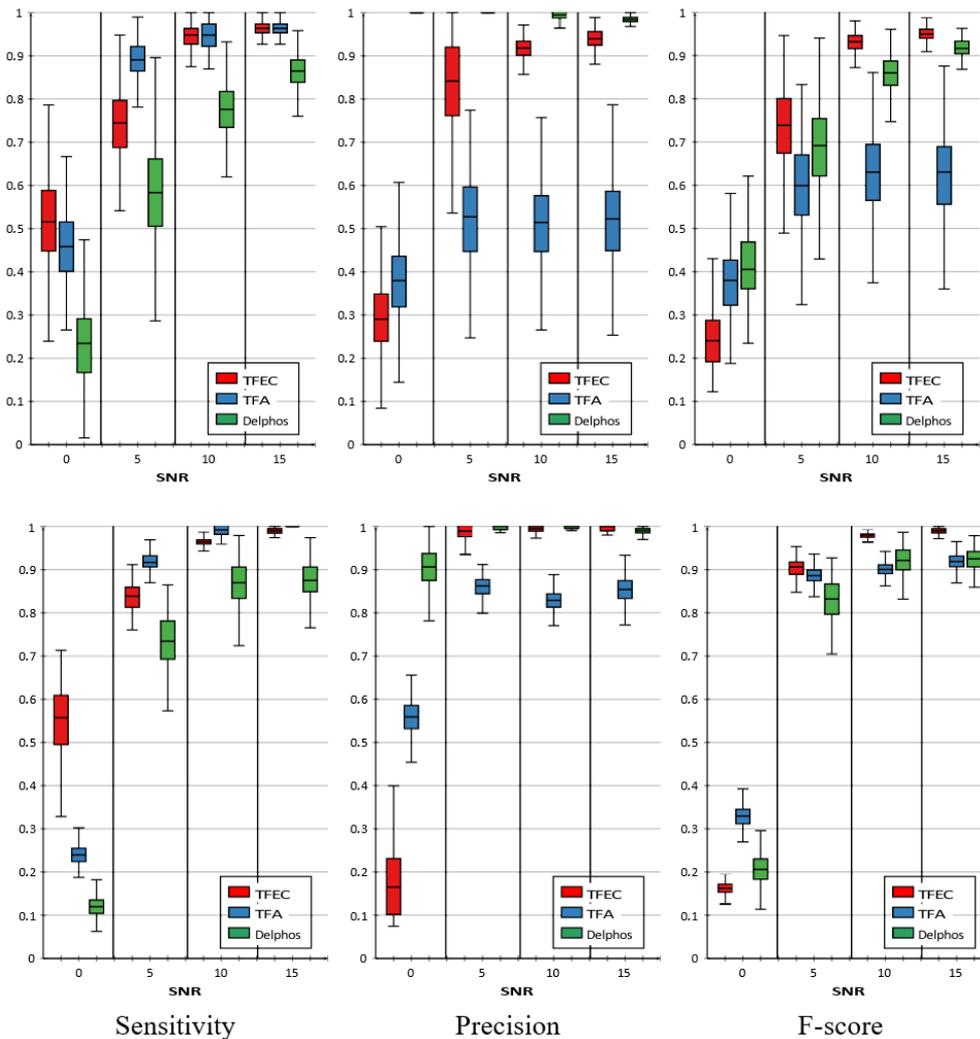

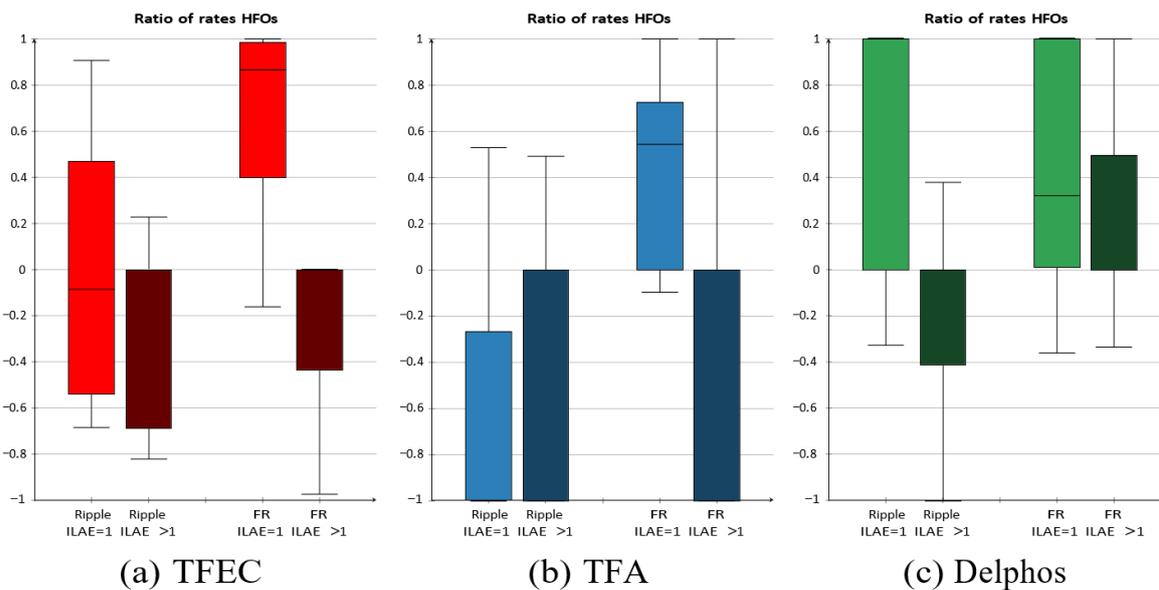

**Figure 5.** Boxplots of detector results in terms of sensitivity, precision, and F-score with varying SNR in ripple and fast ripple bands.

**Figure 6.** Ratio of events in resected and non-resected electrodes in patients with good versus poor outcomes. We observe that fast ripples are more strongly associated with seizure freedom using our method.

166

Figure 7 shows the ratio of HFOs rate for individual patients using our method. Two patients had no HFOs in the FR band (#10 and #18). The plot indicates that, except for patients #3 and #16, all patients with favorable surgical outcomes had the majority of HFOs in the FR band resected (values > 0). This result can be interpreted differently: all patients with most HFOs remaining (values < 0) experienced poor surgical outcomes (except for patients #9 and #17). It should be noted that many channels in patient #17 were removed, yet the patient experienced seizure recurrence.

## 4. DISCUSSION
### 4.1. Contribution

The primary objective of this study was to develop an automatic high-frequency oscillation (HFO) detection method based on time-frequency analysis for extracting events and classifying them as HFO or non-HFO. It was hypothesized that removing HFO contacts would lead to seizure freedom in epileptic patients. To achieve this, we established a straightforward pipeline for HFO detection based on clustering, which does not require any training phase or parameter tuning like classification algorithms. The method can be easily extended for real-time detection of HFOs in epilepsy monitoring units.

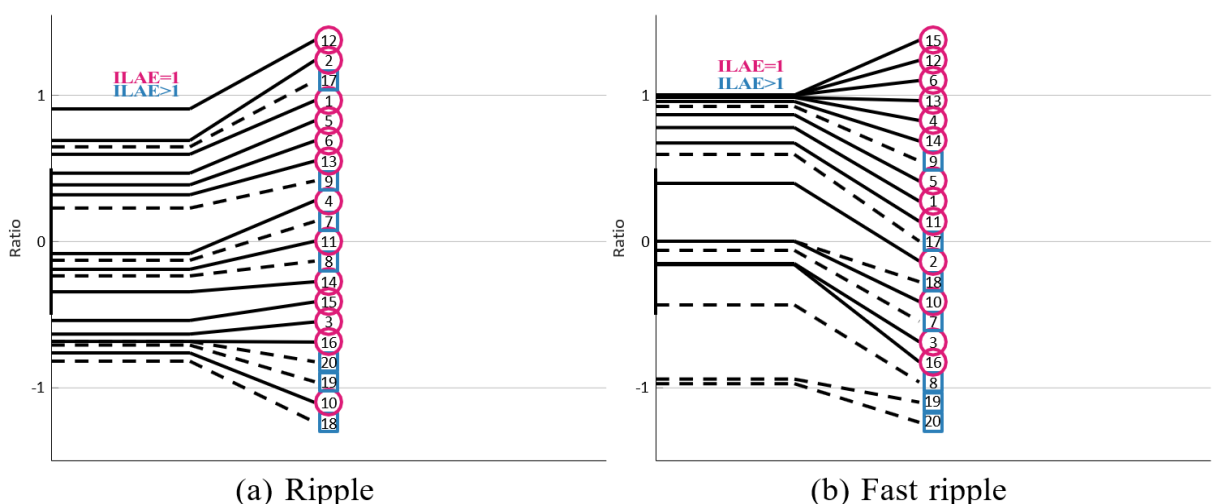

**Figure 7.** The ratio of the HFO rate for all patients.

Note: A good surgery outcome is indicated with a red circle, a poor outcome is indicated with a blue square, and patient numbers are inside the shapes. Ripple events are in column A, and fast ripple is in column B. Ratios of > 0 represent that the majority of that event is respected and the patients have good outcomes, while ratios < 0 represent that the events remain untouched and patients have poor outcomes.

Detectors in the literature require many thresholding parameters for tuning and usually perform well on the studied signal [3, 7]. Our algorithm can be applied easily to different datasets and does not require many parameters for tuning.

Our algorithm uses the S-transform to compute TFD maps, which provide high-resolution maps compared to other time-frequency maps. It achieved an F-score of approximately 98% for high SNR signals at 5, 10, and 15 dB.

By using unsupervised methods, the model can discover the structure of the HFO signal without requiring any prior training. In contrast, supervised methods need labeled data for training, which can introduce bias against untrained data and may not perform well on different datasets. To demonstrate the effectiveness of our method, we validated it against a controlled dataset with varying levels of noise and conducted another experiment with an epileptic patient dataset to correlate with seizure outcomes.

*4.2. Validation with Labelled Dataset*

The performance of the proposed method has been validated against a public dataset and compared with the Delphos detector Liu et al. [10] and Zelmann et al. [7], whose implementations are also publicly available.

While existing detectors in the literature use information in the time or frequency domain only for classifying events into HFO and non-HFOs [3, 6, 7], our method employs joint time and frequency information to extract events from signals, resulting in a significant improvement in HFO detection. It is expected that the F-score will be low in low SNR conditions. According to the definition of HFOs, an event should be salient from background activities in the TFDs map. Therefore, there are few events detected in the first step at SNR 0 dB; consequently, detection accuracy is low.

As can be seen from Figure 5, the results of this study indicate that our method outperforms existing methods at different SNRs, but compared to Delphos, it shows lower performance at low SNR 0. This is because the Delphos method uses a different pre-whitening technique in its initial processing step. In terms of F-score, our method performed better than the Delphos and TFA methods at high SNRs of 5, 10, and 15 dB.

*4.3. Validation with Epileptic Patients Dataset*

Our algorithm is applied to real data obtained from epileptic patients with follow-up surgery outcomes. While in the controlled dataset, we could compare the algorithm results with labeled events, here we can examine the relationship between HFO rate and surgical outcomes.

We hypothesized that patients in whom the majority of HFOs are removed have good surgical outcomes, and patients in whom HFOs remain post-surgically have poor outcomes. As shown in Figure 7, the most important clinically relevant finding was that HFOs generating tissue are highly linked to achieving seizure freedom. In patients with a good outcome, the majority of tissue-generating HFOs (ripple and fast-ripple) events are resected (except patients #9 and #17). Conversely, in all patients with poor outcomes (except patients #3 and #16 in the fast-ripple band), epileptogenic tissue was not resected. Although it has been shown that fast ripples are more strongly linked to epileptogenicity than ripples in Fedele et al. [13], and their detector (TFA), here we show that both ripples and fast ripples are related to surgical outcomes by our and Delphos methods (ratios > 0). Nonetheless, fast ripples are more associated with epileptogenicity than ripples. According to Figure 6, by applying detectors on all patients, we can conclude that both ripples and fast ripples are indicators of epileptogenicity, but fast ripples are more strongly linked because the ratio of the rate in fast ripples is higher than that of ripples. Ripples and fast ripples exhibit different characteristics.

Distribution patterns, while fast ripples are more localized than ripples. Therefore, interpreting HFO rates as a marker of epileptogenicity requires differentiating between ripples and fast ripples. While fast ripples show a strong link to the epileptogenic zone, the distribution of ripples is less indicative of the epileptogenic zone. These results are consistent with those of other studies [4, 8] and suggest that there is a strong correlation between removed HFO contacts (especially fast-ripple contacts) and surgical outcomes.

*4.4. Clinical Implications*

HFOs appear to be a reliable biomarker of epileptogenicity in patients with epilepsy. Patients whose HFO areas overlap more with the resected area tend to have better surgical outcomes. Conversely, patients whose HFO areas remain untouched often experience poorer outcomes. Fast ripples correlate more strongly with seizure outcomes than ripples and are more specific to epileptogenicity.

## 5. CONCLUSIONS

High-frequency oscillations are valuable biomarkers for epilepsy and are used to delineate the epileptogenic zone. The current study aimed to develop an HFO detector based on unsupervised learning, which identifies blobs in time-frequency domain maps and clusters them into HFO and non-HFO. We tested our method on two public datasets, involving controlled and epileptic patients, to evaluate the detector's performance. Our findings were consistent with previous research and indicated that fast ripple could be a promising biomarker of epileptogenicity in epileptic patients. We found a significant correlation between the removal of the HFO area, especially fast ripple, and seizure freedom and recurrence in the remaining HFO area.

**Funding:** This study received no specific financial support.
**Institutional Review Board Statement:** Not applicable.
**Transparency:** The authors state that the manuscript is honest, truthful, and transparent, that no key aspects of the investigation have been omitted, and that any differences from the study as planned have been clarified. This study followed all writing ethics.
**Competing Interests:** The authors declare that they have no competing interests.
**Authors' Contributions:** All authors contributed equally to the conception and design of the study. All authors have read and agreed to the published version of the manuscript.

## REFERENCES

[1] M. Navarrete, C. Alvarado-Rojas, M. Le Van Quyen, and M. Valderrama, "RIPPLELAB: A comprehensive application for the detection, analysis and classification of high frequency oscillations in electroencephalographic signals," *PloS One*, vol. 11, no. 6, p. e0158276, 2016. https://doi.org/10.1371/journal.pone.0158276

[2] J. Jacobs, P. LeVan, R. Chander, J. Hall, F. Dubeau, and J. Gotman, "Interictal high-frequency oscillations (80–500 Hz) are an indicator of seizure onset areas independent of spikes in the human epileptic brain," *Epilepsia*, vol. 49, no. 11, pp. 1893-1907, 2008. https://doi.org/10.1111/j.1528-1167.2008.01656.x

[3] R. J. Staba, C. L. Wilson, A. Bragin, I. Fried, and J. Engel Jr, "Quantitative analysis of high-frequency oscillations (80–500 Hz) recorded in human epileptic hippocampus and entorhinal cortex," *Journal of Neurophysiology*, vol. 88, no. 4, pp. 1743-1752, 2002. https://doi.org/10.1152/jn.2002.88.4.1743

[4] J. Jacobs *et al.*, "High-frequency electroencephalographic oscillations correlate with outcome of epilepsy surgery," *Annals of Neurology: Official Journal of the American Neurological Association and the Child Neurology Society*, vol. 67, no. 2, pp. 209-220, 2010. https://doi.org/10.1002/ana.21847

[5] A. B. Gardner, G. A. Worrell, E. Marsh, D. Dlugos, and B. Litt, "Human and automated detection of high-frequency oscillations in clinical intracranial EEG recordings," *Clinical Neurophysiology*, vol. 118, no. 5, pp. 1134-1143, 2007. https://doi.org/10.1016/j.clinph.2006.12.019

[6] B. Crépon *et al.*, "Mapping interictal oscillations greater than 200 Hz recorded with intracranial macroelectrodes in human epilepsy," *Brain*, vol. 133, no. 1, pp. 33-45, 2010. https://doi.org/10.1093/brain/awp277

[7] R. Zelmann, F. Mari, J. Jacobs, M. Zijlmans, F. Dubeau, and J. Gotman, "A comparison between detectors of high frequency oscillations," *Clinical Neurophysiology*, vol. 123, no. 1, pp. 106-116, 2012. https://doi.org/10.1016/j.clinph.2011.06.006

[8] T. Fedele *et al.*, "Automatic detection of high frequency oscillations during epilepsy surgery predicts seizure outcome," *Clinical Neurophysiology*, vol. 127, no. 9, pp. 3066-3074, 2016. https://doi.org/10.1016/j.clinph.2016.06.009

[9] S. Burnos, B. Frauscher, R. Zelmann, C. Haegelen, J. Sarnthein, and J. Gotman, "The morphology of high frequency oscillations (HFO) does not improve delineating the epileptogenic zone," *Clinical Neurophysiology*, vol. 127, no. 4, pp. 2140-2148, 2016. https://doi.org/10.1016/j.clinph.2016.01.002

[10] S. Liu *et al.*, "Stereotyped high-frequency oscillations discriminate seizure onset zones and critical functional cortex in focal epilepsy," *Brain*, vol. 141, no. 3, pp. 713-730, 2018. https://doi.org/10.1093/brain/awx374

[11] C. Migliorelli *et al.*, "SGM: A novel time-frequency algorithm based on unsupervised learning improves high-frequency oscillation detection in epilepsy," *Journal of Neural Engineering*, vol. 17, no. 2, p. 026032, 2020.

[12] N. Roehri, J.-M. Lina, J. C. Mosher, F. Bartolomei, and C.-G. Bénar, "Time-frequency strategies for increasing high-frequency oscillation detectability in intracerebral EEG," *IEEE Transactions on Biomedical Engineering*, vol. 63, no. 12, pp. 2595-2606, 2016.

[13]   T. Fedele *et al.*, *High frequency oscillations detected in the intracranial EEG of epilepsy patients during interictal sleep, patients' electrode location and outcome of epilepsy surgery*. Berkeley, CA: CRCNS.org, University of California, Berkeley, 2017.

[14]   R. G. Stockwell, L. Mansinha, and R. Lowe, "Localization of the complex spectrum: The S transform," *IEEE Transactions on Signal Processing*, vol. 44, no. 4, pp. 998-1001, 1996.

[15]   N. Otsu, "A threshold selection method from gray-level histograms," *IEEE Transactions on Systems, Man, and Cybernetics*, vol. 9, no. 1, pp. 62–66, 1979.

[16]   B. Boashash and S. Ouelha, "Designing high-resolution time–frequency and time–scale distributions for the analysis and classification of non-stationary signals: A tutorial review with a comparison of features performance," *Digital Signal Processing*, vol. 77, pp. 120-152, 2018. https://doi.org/10.1016/j.dsp.2017.07.015

[17]   B. Boashash, H. Barki, and S. Ouelha, "Performance evaluation of time-frequency image feature sets for improved classification and analysis of non-stationary signals: Application to newborn EEG seizure detection," *Knowledge-Based Systems*, vol. 132, pp. 188-203, 2017. https://doi.org/10.1016/j.knosys.2017.06.015

[18]   M. Z. Gashti, "Detection of spam email by combining harmony search algorithm and decision tree," *Engineering, Technology & Applied Science Research*, vol. 7, no. 3, pp. 1713-1718, 2017. https://doi.org/10.48084/etasr.1171

[19]   N. Roehri, F. Pizzo, F. Bartolomei, F. Wendling, and C.-G. Bénar, "What are the assets and weaknesses of HFO detectors? A benchmark framework based on realistic simulations," *PloS one*, vol. 12, no. 4, p. e0174702, 2017. https://doi.org/10.1371/journal.pone.0174702

[20]   S. Burnos *et al.*, "Human intracranial high frequency oscillations (HFOs) detected by automatic time-frequency analysis," *PloS one*, vol. 9, no. 4, p. e94381, 2014. https://doi.org/10.1371/journal.pone.0094381